\documentclass[sigconf]{acmart}
\renewcommand\footnotetextcopyrightpermission[1]{}

\usepackage{multirow}
\usepackage{array}
\usepackage{float}

\AtBeginDocument{%
  \providecommand\BibTeX{{%
    \normalfont B\kern-0.5em{\scshape i\kern-0.25em b}\kern-0.8em\TeX}}}

\begin{document}

\title{FantasyTalking: Realistic Talking Portrait Generation via Coherent Motion Synthesis}


\author{Mengchao Wang}
\authornote{Equal contribution}
\affiliation{%
  \institution{AMAP, Alibaba Group}
  }
\email{wangmengchao.wmc@alibaba-inc.com}

\author{Qiang Wang}
\authornotemark[1]
\affiliation{
  \institution{AMAP, Alibaba Group}
  }
\email{yijing.wq@alibaba-inc.com}

\author{Fan Jiang}
\authornote{Project leader}
\affiliation{
  \institution{AMAP, Alibaba Group}
  }
\email{frank.jf@alibaba-inc.com}

\author{Yaqi Fan}
\affiliation{
  \institution{Beijing University of Posts and Telecommunications}
  }
\email{yqfan@bupt.edu.cn}

\author{Yunpeng Zhang}
\affiliation{
  \institution{AMAP, Alibaba Group}
  }
\email{daoshi.zyp@alibaba-inc.com}

\author{Yonggang Qi}
\authornote{Corresponding author}
\affiliation{
  \institution{Beijing University of Posts and Telecommunications}
  }
\email{qiyg@bupt.edu.cn}

\author{Kun Zhao}
\affiliation{
  \institution{AMAP, Alibaba Group}
  }
\email{kunkun.zk@alibaba-inc.com}

\author{Mu Xu}
\affiliation{
  \institution{AMAP, Alibaba Group}
  }
\email{xumu.xm@alibaba-inc.com}


\begin{abstract}
Creating a realistic animatable avatar from a single static portrait remains challenging. Existing approaches often struggle to capture subtle facial expressions, the associated global body movements, and the dynamic background. To address these limitations, we propose a novel framework that leverages a pretrained video diffusion transformer model to generate high-fidelity, coherent talking portraits with controllable motion dynamics. At the core of our work is a dual-stage audio-visual alignment strategy. In the first stage, we employ a clip-level training scheme to establish coherent global motion by aligning audio-driven dynamics across the entire scene, including the reference portrait, contextual objects, and background. In the second stage, we refine lip movements at the frame level using a lip-tracing mask, ensuring precise synchronization with audio signals. To preserve identity without compromising motion flexibility, we replace the commonly used reference network with a facial-focused cross-attention module that effectively maintains facial consistency throughout the video. Furthermore, we integrate a motion intensity modulation module that explicitly controls expression and body motion intensity, enabling controllable manipulation of portrait movements beyond mere lip motion. Extensive experimental results show that our proposed approach achieves higher quality with better realism, coherence, motion intensity, and identity preservation. Ours project page: \url{https://fantasy-amap.github.io/fantasy-talking/}.
\end{abstract}





\maketitle

\begin{figure*}[h]
  \includegraphics[width=\linewidth]{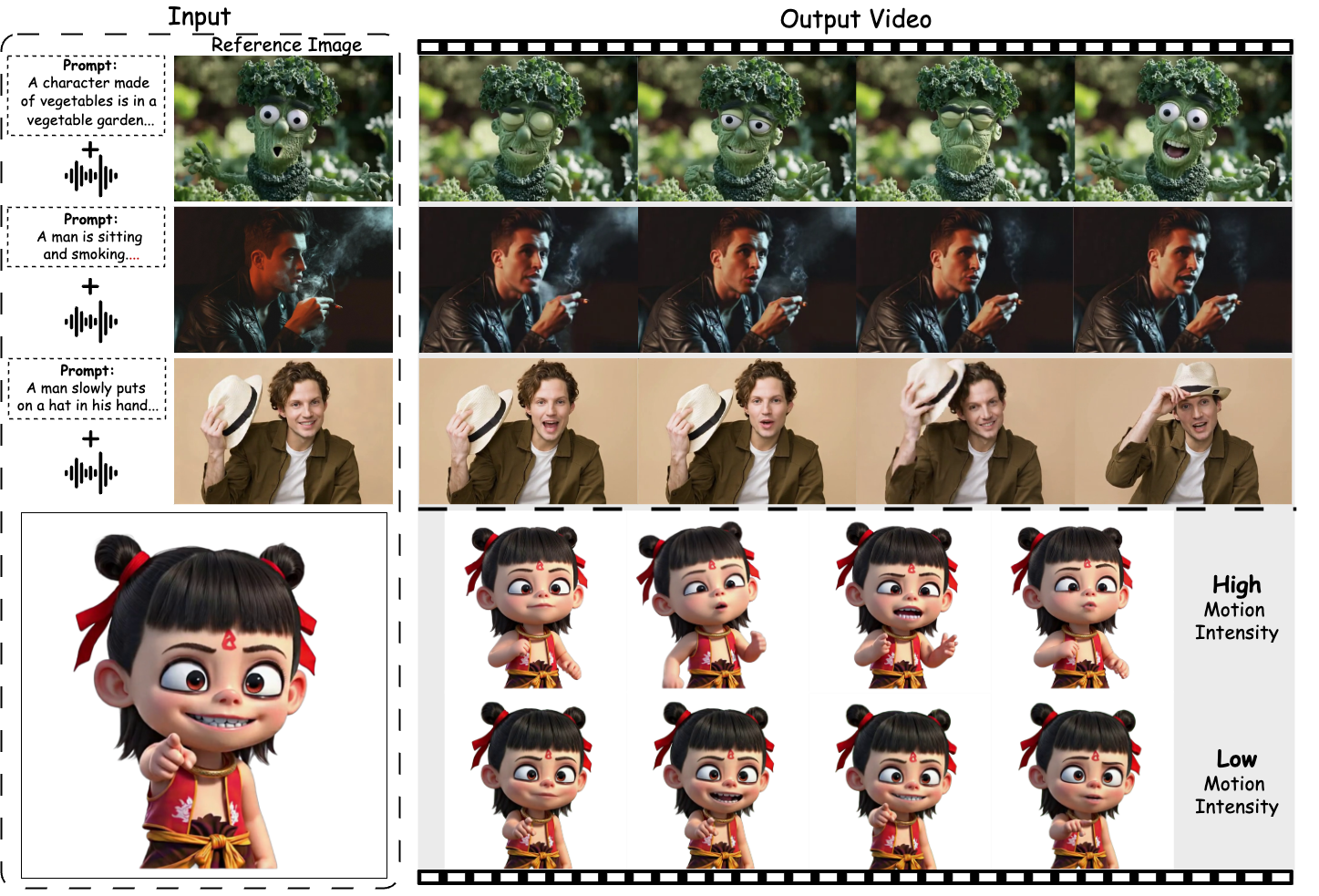}
  \caption{Given a portrait image, voice and text, FantasyTalking can generate animated portraits with rich expressions, natural body movements, and identity features. In addition, FantasyTalking can control the motion intensity of animated portraits. Please refer to our supplementary materials for the video results.}
  \label{fig:fig0}
\end{figure*}

\section{Introduction}
Generating an animatable avatar from a single static portrait image has long been a fundamental challenge in computer vision and graphics. In particular, the ability to synthesize a realistic talking avatar given a reference image unlocks a wide range of applications in gaming, filmmaking, and virtual reality. It is crucial that the avatar can be seamlessly controlled using audio signals, enabling intuitive and flexible manipulation of expressions, lip movements, and gestures to align with the desired content. 

Early attempts \cite{song2022audio, guan2023stylesync, chatziagapi2023lipnerf, zhang2023sadtalker, ma2023dreamtalk, wei2024aniportrait} to tackle this task mainly resort to 3D intermediate representations, such as 3D Morphable Models (3DMM) \cite{tran2018nonlinear} or FLAME \cite{li2017learning}. However, these approaches typically face challenges in accurately capturing subtle expressions and realistic motions, which significantly limits the quality of the generated portrait animations. Recent research \cite{xu2024hallo, tian2024emo, cui2024hallo2, jiang2024loopy, chen2024echomimic} has increasingly focused on creating talking head videos using diffusion models, which show great promise in generating visually compelling content that adheres to multi-modal conditions, such as reference images, text prompts, and audio signals. However, the realism of the generated videos remains unsatisfactory. Existing methods typically focus on tame talking head scenarios, achieving precise audio-aligned lip movements while neglecting other related motions, such as facial expressions and body movements, both of which are essential for producing smooth and coherent portrait animations. Moreover, the background and contextual objects usually remain static throughout the animation, which makes the scene less natural. 

In this work, we leverage pretrained video diffusion transformer models to generate highly realistic and visually coherent talking portraits. In essence, we propose a multi-modal alignment framework built on the DiT-based video generation model to encourage unified dynamics across the whole scene, encompassing the reference portrait, associated contextual objects, and the background. Technically, we propose a dual-stage audio-visual alignment strategy to facilitate portrait video generation. In the first stage, leveraging the powerful temporospatial modeling capabilities of the DiT-based model, we devise a clip-level training to capture diverse implicit connections between the audio and visual dynamics across the entire clip. This enables an overall coherent generation of global motion.  Lip movements are critical for enhancing the quality of the portrait video. However, the lip typically only occupies a small region in a frame, so it is challenging to precisely align lip movements with the audio signals on the entire frame. Therefore, in the second stage, we learn the attention of visual tokens mapped from audio tokens and employ a mask that enforces the refinement of lip movements, ensuring they adhere more closely to the audio content at the frame level. Moreover, we avoid using the commonly adopted reference network for identity preservation. We found out that such an approach typically references the entire image and severely restricts the dynamic effects of the portrait. Instead, we reveal that a cross-attention module focusing on facial modeling effectively ensures identity consistency throughout the video. Lastly, we introduce a motion intensity conditioning module that decouples the character's expressions and body movements, thereby enabling the manipulation of motion intensity in the generated dynamic portrait. 

In summary, our contributions are as follows:
\begin{itemize}
    \item We devise a dual-stage audio-visual alignment training strategy to adapt a pretrained video generation model to first establish coherent global motions involving background and contextual objects other than the portrait itself, corresponding to input audio at clip level, then construct precisely aligned lip movements to further improve the quality of the generated video. 
    \item Instead of adopting the conventional reference network for identity preservation, we streamline the process by devising a facial-focused cross-attention module that concentrates on modeling facial regions and guides the video generation with consistent identity.
    \item We integrate a motion intensity modulation module that explicitly controls facial expression and body motion intensity, enabling controllable manipulation of portrait movements beyond mere lip motion.
    \item Extensive experiments demonstrate that our proposed approach achieves new SOTA in terms of video quality, temporal consistency, and motion diversity.
\end{itemize}

\begin{figure*}[h]
  \includegraphics[width=\linewidth]{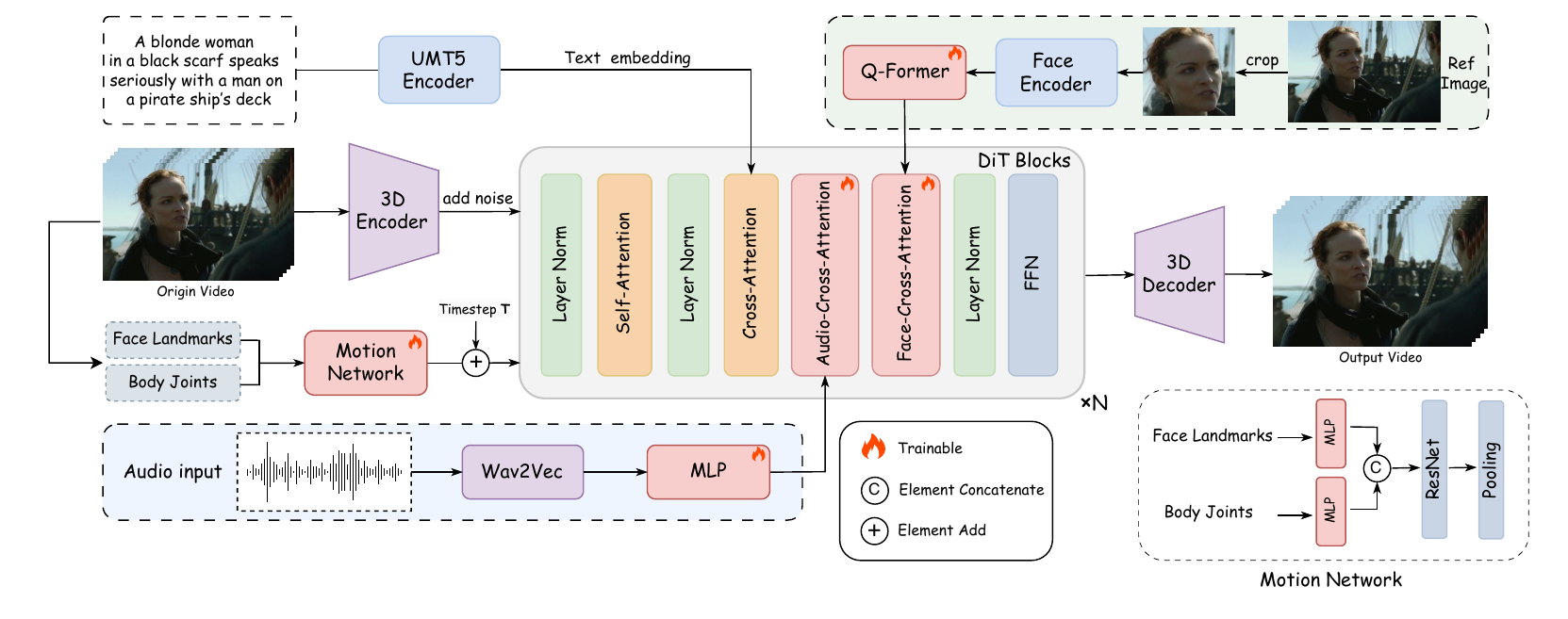}
  \caption{Overview of FantasyTalking.}
  \label{fig:overview}
\end{figure*}

\section{Related Work}
\subsection{Diffusion-Based Video Generation}
The remarkable achievements of diffusion models in image generation \cite{dhariwal2021diffusion, esser2024scaling, rombach2022high} have inspired extensive research into video generation \cite{singer2022make, kong2024hunyuanvideo, ho2022video}. Early methods employing diffusion models predominantly relied on the UNet architecture\cite{ronneberger2015u}, with notable examples being AnimateDiff \cite{guo2023animatediff} and Stable Video Diffusion \cite{blattmann2023stable}. These approaches, leveraging pretrained image generation models, harness their robust spatial generation capabilities and incorporate specifically designed temporal layers to acquire motion-related understanding. More recently, models based on the DiT architecture \cite{peebles2023scalable} have significantly propelled the advancement of video generation technology \cite{yang2024cogvideox, kong2024hunyuanvideo, wan2.1, xu2024easyanimate}. These models employ 3D VAE \cite{kingma2013auto} as the encoder and decoder, coupled with the Transformer’s formidable sequence modeling prowess, showcasing substantial potential in tackling intricate video generation tasks. They have demonstrated impressive capabilities in maintaining human identity \cite{yuan2024identity, zhang2025fantasyid}, controlling expressions \cite{qiu2025skyreels}, and virtual try-on \cite{zheng2024dynamic} applications, among others. 

\subsection{Audio-driven Talking head Generation}
The task of synthesizing realistic talking face videos from input audio has remained a persistent research focus. Early approaches \cite{zhang2023sadtalker, ma2023dreamtalk, wei2024aniportrait} employed 3D intermediate representations, utilizing facial animation parameters derived from 3D Morphable Models (3DMM)  as guidance for video generation. However, the limited expressiveness of 3DMM in capturing intricate facial expressions and head movements significantly constrained the authenticity and naturalness of synthesized videos. In contrast, emerging end-to-end audio-to-video synthesis methods \cite{tian2024emo, jiang2024loopy, chen2024echomimic, cui2024hallo3} demonstrate enhanced potential, yet still face two critical challenges. Firstly, existing approaches typically employ reference networks initialized from backbone architectures to preserve speaker identity, and the input of the reference network is the whole image rather than focusing on the face, which inadvertently restricts the model's capacity to generate videos with broader motion ranges. Secondly, although prior methods have emphasized precise audio-lip synchronization, the inherent weak correlations between audio signals and other facial expressions and body movements remain largely underexplored. Despite Hallo3 initially progress in the wild talking head task, the areas of facial-focused identity preservation and complex scene interaction are yet to be thoroughly explored.

\section{Method}

Given a sigle reference image, a driving audio and a prompt, FantasyTalking is designed to generate the video synchronized with the audio while ensuring that the identity characteristics of the person are maintained during their actions. An overview of FantasyTalking is illustrated in Figure \ref{fig:overview}. We investigate a Dual-Stage method to maintain audio-to-visual alignment when injecting audio signals (Sec. \ref{subsec:dava}). Additionally, we employ an identity learning method to preserve the identity characteristics in the video (Sec. \ref{subsec:indentity}) and a motion network to control the expressions and the motion intensity (Sec. \ref{subsec:motion}). The following section (Sec. \ref{subsec:Preliminaries}) elaborates on the preliminaries of our method. 

\subsection{Preliminaries} \label{subsec:Preliminaries}
\noindent {\bf{Latent Diffusion Model.}}~ Our method is built upon the Latent Diffusion Model (LDM) , which is a framework that learns in the latent space rather than the pixel space. During training, we use a pre-trained VAE encoder $E$ to compress video data $x$ from the pixel space into latent tokens $z = E(x)$. During training, the Gaussian noise $\epsilon$ is progressively added to $z$ to create $z_t = \sqrt{\alpha_t} z + \sqrt{1 - \alpha_t} \epsilon$ at $t$ timestep. Here, $\alpha_t$ represents as the noise scheduler. The training objective of the LDM focuses on a reconstruction loss that aims to minimize the difference between the added noise and the noise predicted by the network $\epsilon_\theta$: 

\begin{equation}
    L = \mathbb{E}_{t, z_t, c, \epsilon \sim \mathcal{N}(0, 1)} \left[\|\epsilon_\theta(\mathbf{z}_t, t, c) - \epsilon\|_2^2\right]
    \label{eq:loss1}
\end{equation}

where $c$ denotes the conditions like audio, text or images. In the inference phase, the model iteratively denoises latent sampled from a Gaussian distribution. Subsequently, the denoised latent representations are decoded back into videos using the VAE decoder $D$.

\noindent {\bf{Diffusion Transformer.}}~ The Diffusion Transformer (DiT) \cite{peebles2023scalable} is a diffusion model designed based on the Transformer architecture \cite{vaswani2017attention}, showcasing significant potential in the field of video generation. Specifically, we adopt Wan2.1 \cite{wan2.1} as the foundational architecture. This model employs a causal 3D VAE to compress videos both temporally and spatially, while utilizing UMT5 \cite{chung2023unimax} to encode textual information, yielding the text-conditioned input $c_{text}$. The text embeddings are then integrated into the DiT through cross-attention mechanisms. In addition, the embeddings of the timestep $t$ are injected into the model by predicting six modulation parameters individually.

\subsection{Dual-Stage Audio-Visual Alignment} \label{subsec:dava}
\begin{figure}[h]
  \includegraphics[width=\linewidth]{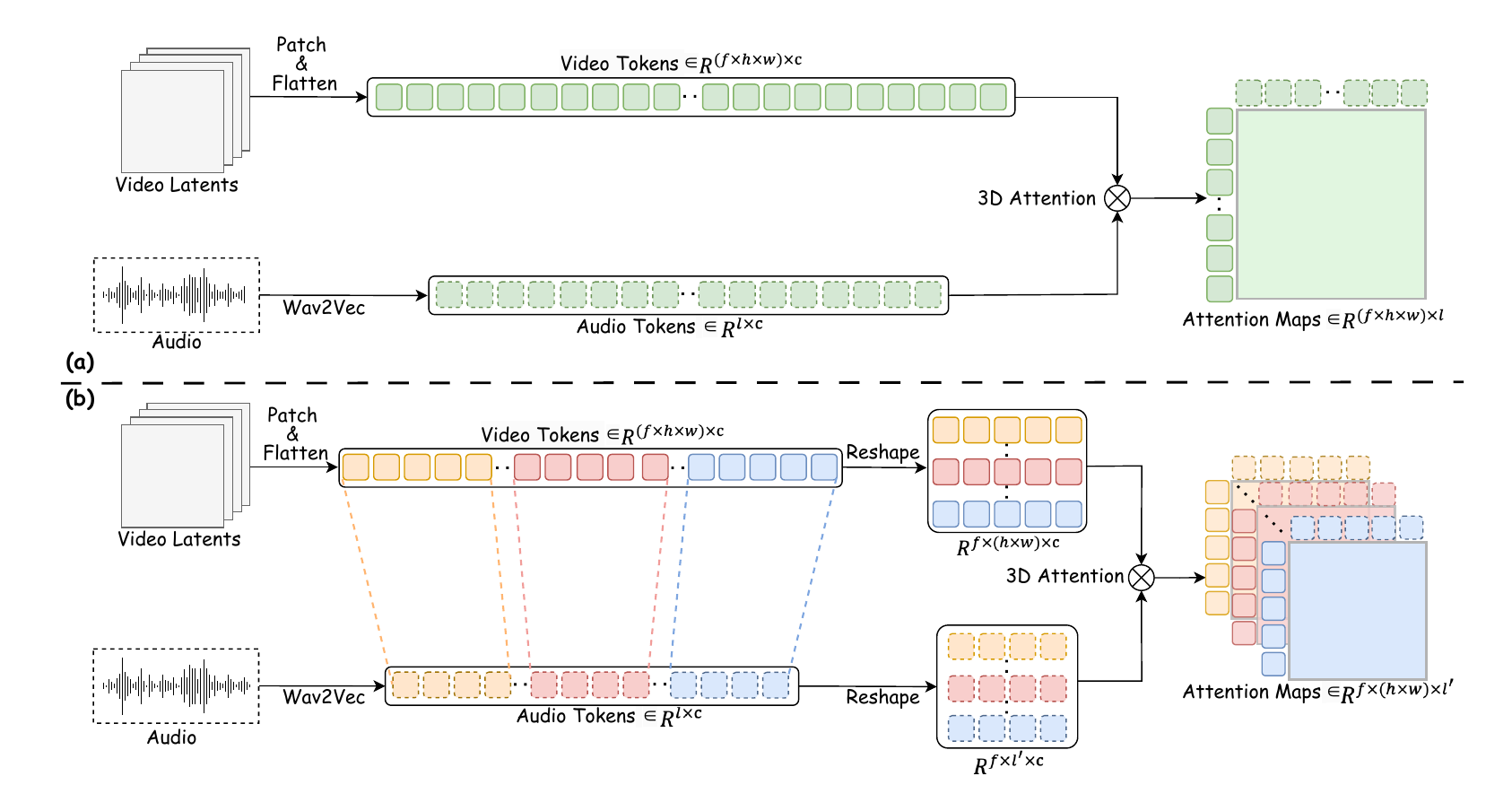}
  \caption{Dual-Stage Audio-Visual Alignment.}
  \label{fig:dava}
\end{figure}

\noindent {\bf{Audio-Visual Alignment.}}~ We utilize Wav2Vec \cite{schneider2019wav2vec} to extract audio tokens containing multi-scale rich acoustic features. As shown in Figure \ref{fig:dava}, the audio tokens length $l$ differs from that of the video tokens length $(f \times h \times w)$, where $f$, $h$ and $w$ are the frame numbers, height and width of latent videos. There exists a one-to-one mapping relationship between these two token sequences. The task of tame talking head video generation typically focuses on the frame-level alignment of lip movements. However, wild talking head generation requires attention not only to the lip movements that are directly correlated with the audio but also to the movements of other facial components and body parts that are weakly correlated with the audio features, such as eyebrows, eyes, and shoulders. These movements are not strictly temporally aligned with the audio. To address this, we propose a Dual-Stage Audio-Vision Alignment approach. In the first training stage, we learn visual features related to the audio at the clip level. In the second training stage, we focus on the visual features that are highly correlated with the audio at the frame level. 

\noindent {\bf{Clip-Level Training.}}~ As illustrated in Figure \ref{fig:dava}(a), the first training stage computes 3D full attention correlations across full-length audio-visual token sequences at the clip level, establishing global audiovisual dependencies while enabling holistic feature fusion. While this stage enables joint learning of both weakly audio-correlated non-verbal cues (e.g., eyebrow movements, shoulder motions) and strongly audio-synchronized lip dynamics, but the model struggles to learn precise lip movements. This is due to the fact that the lips occupy only a small portion of the entire visual field, while the video sequence is highly correlated with the audio in each frame. 

\noindent {\bf{Frame-Level Training.}}~ In the second training stage, as depicted in Figure \ref{fig:dava}(b), we focus exclusively on lip-centric motion refinement through frame-exact audio-visual alignment. We segment the audio and videos according to a one-to-one mapping relationship, reshape the video tokens into the shape of $f \times (h \times w) \times c$ and the audio tokens into the shape of $f \times l' \times c$, where $c$ represents the number of channels. Subsequently, we compute the 3D full attention between these tokens, ensuring that the visual features attend only to their corresponding audio features.

Additionally, in order to focus the attention on the lip area, we leverage MediaPipe \cite{lugaresi2019mediapipe} to extract precise lip masks in pixel space, which are then projected into the latent space via trilinear interpolation, forming our lip-focused constraint mask $M$. The frame-level loss in Eq. \ref{eq:loss1} is thus reweighted as: 

\begin{equation}
    {L}_c = M \odot {L}
    \label{eq:loss2}
\end{equation}

where $\odot$ denotes element-wise multiplication. However, exclusive reliance on lip-specific constraints risks over-regularization, suppressing natural head movements and background dynamics. To mitigate this issue, we employ a probability $\eta$ to control the application of the constraint, allowing the model to balance between focusing on lip movements and maintaining the naturalness of overall movements. 

\begin{equation}
    L' = 
    \begin{cases}
    L_c, & \text{if } p > \eta \\
    L, & \text{otherwise}
    \end{cases}
    \label{eq:loss3}
\end{equation}

\subsection{Identity Preservation} \label{subsec:indentity}
While audio conditioning effectively establishes correlations between acoustic inputs and character motions, prolonged video sequences and intensified movements often lead to rapid identity degradation in synthesized results. Previous methods \cite{tian2024emo, jiang2024loopy, chen2024echomimic, cui2024hallo3} typically employ reference networks initialized from the backbone model to preserve identity characteristics, yet these methods exhibit two critical limitations. Firstly, the reference network processes full-frame images rather than facial regions of interest, biasing the model towards generating static backgrounds and motions with constrained expressiveness. Secondly, the reference network model typically has a network structure similar to that of the backbone model, resulting in a high degree of redundancy in their feature representation capabilities, and increases the computational load and complexity of the model. 

To address this issue, we propose an identity preservation method to maintain consistency of facial features. Specifically, we first crop the facial region from the reference image \cite{deng2019accurate} to ensure that the model only focuses on identity related facial regions. Subsequently, we utilize ArcFace \cite{deng2019arcface} to extract the facial feature and then employ Q-Former \cite{li2023blip} for alignment, resulting in the ID embedding $F_{id}$. Similar to audio conditioning, these identity features interact with each pretrained DiT attention block through dedicated cross-attention layers. Formally, the hidden state $Z_i$ of each DiT block is reformulated as:

\begin{equation}
    Z_i' = Z_i + \lambda_1 * \text{Attention}(Q_i, K_i^a, V_i^a) + \lambda_2 * \text{Attention}(Q_i, K_i^{id}, V_i^{id})
\end{equation}

where $i$ represents the layer number of the attention block, $Q_i$ is query matrices, $K_i^a$ and $K_i^{id}$ are the audio and identity key matrices, $V_i^a$ and $V_i^{id}$ are the audio and identity values matrices of the attention operation. The hyperparameters $\lambda_1$ and $\lambda_2$ control the relative contributions of audio and identity conditioning.

\subsection{Motion Intensity Modulation Network} \label{subsec:motion}

Individual speaking styles exhibit significant variations in facial expressions and body movement amplitudes, which cannot be explicitly controlled solely through audio and identity conditioning. Particularly in the context of wild talking head scenarios, the character's expressions and body movements are more varied and dynamic compared to tame talking head scenarios. Therefore, we introduce a motion intensity modulation network to govern these dynamics.

Specifically, we utilize Mediapipe \cite{lugaresi2019mediapipe} to extract the variance of facial landmark keypoint sequences, denoted as facial expression movement coefficient $\omega_l$, and DWPose \cite{yang2023effective} to compute the variance of body joint sequences, denoted as body movement cofficient $\omega_b$. Both $\omega_l$ and $\omega_b$ are normalized to the range [0, 1], representing the intensity of facial expressions and body movements, respectively. As illustrated in Figure \ref{fig:overview}, motion intensity modulation network consists of MLP layers, a ResNet layer \cite{he2016deep}, and an average pooling layer. The resulting motion embeddings are added with the timesteps. During inference stage, users are allowed to customize the input coefficient $\omega_l$ and $\omega_b$ to control the amplitude of facial and body motion intensity.

\begin{table*}
    \begin{tabular}{@{}ll|c|c|c|c|c|c|c|c|c@{}}
    \toprule
    \textbf{Dataset} & \textbf{Method} & \textbf{FID$\downarrow$} & \textbf{FVD$\downarrow$} & \textbf{Sync-C$\uparrow$} & \textbf{Sync-D$\downarrow$} & \textbf{ES$\uparrow$} & \textbf{IDC$\uparrow$} & \textbf{SD$\uparrow$} & \textbf{BD$\uparrow$} & \textbf{Aesthetic$\uparrow$} \\ \midrule
    \multirow{5}{*}{Tame Talking} & Aniportrait & 37.672 & 397.114 & 1.095 & 12.461 & 0.9508 & 0.9372 & 4.639 & - & 0.5129 \\
    & EchoMimic & 33.765 & 471.452 & 2.514 & 10.743 & 0.9527 & 0.9419 & 5.783 & - & 0.5108 \\
    & Sonic & 30.396 & 358.023 & 4.197 & \textbf{9.103} & 0.9595 & 0.9885 & 8.832 & - & 0.5312 \\
    & Hallo3 & 32.617 & 347.358 & 4.060 & 9.371 & 0.9566 & 0.9774 & 8.415 & - & 0.5247 \\
    & FantasyTalking & \textbf{27.695} & \textbf{301.173} & \textbf{4.226} & 9.251 & \textbf{0.9612} & \textbf{0.9892} & \textbf{11.745} & - & \textbf{0.5362} \\ \midrule
    \multirow{5}{*}{Wild Talking} & Aniportrait & 63.574 & 841.962 & 0.996 & 12.084 & 0.9318 & 0.9031 & 2.252 & 1.9287 & 0.5357 \\
    & EchoMimic & 59.746 & 590.373 & 1.949 & 10.754 & 0.9463 & 0.9202 & 3.201 & 1.9508 & 0.5311 \\
    & Sonic & 45.400 & 489.985 & 2.689 & 10.194 & 0.9539 & 0.9607 & 10.484 & 3.9019 & 0.5913 \\
    & Hallo3 & 47.403 & 488.499 & 2.673 & 10.292 & 0.9420 & 0.9538 & 11.411 & 5.2840 & 0.5842 \\
    & FantasyTalking & \textbf{43.137} & \textbf{483.108} & \textbf{3.154} & \textbf{9.689} & \textbf{0.9589} & \textbf{0.9754} & \textbf{13.783} & \textbf{7.9624} & \textbf{0.6183} \\ \bottomrule
    \end{tabular}
    \caption{Comparison of different methods on tame and wild talking head datasets. The
best results are highlighted in bold.}
    \label{tab:comparison}
\end{table*}

\begin{figure*}[h]
  \includegraphics[width=\linewidth]{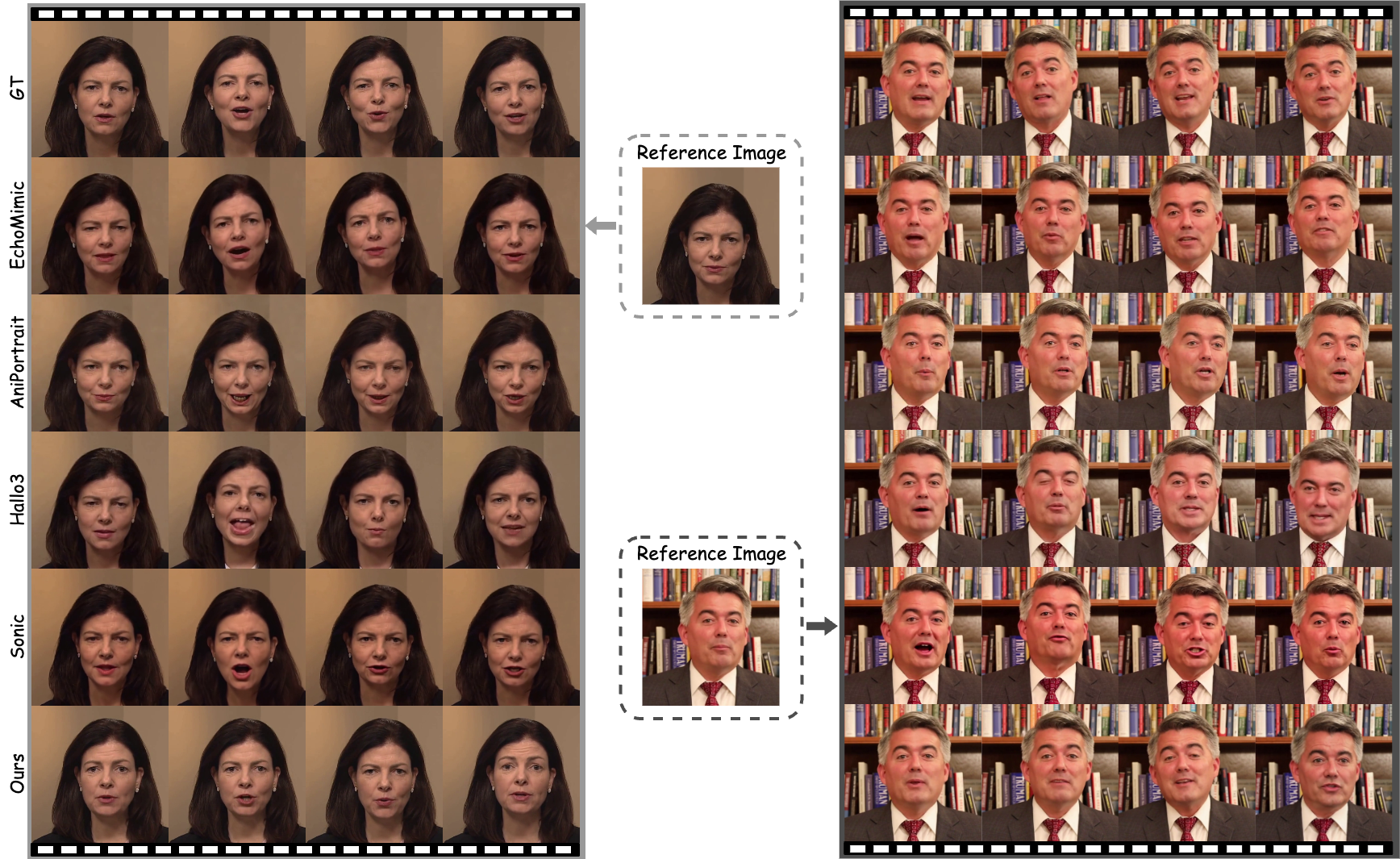}
  \caption{Qualitative comparison on tame talking head dataset (HDTF).}
  \label{fig:compare_hdtf}
\end{figure*}

\begin{figure*}[h]
  \includegraphics[width=\linewidth]{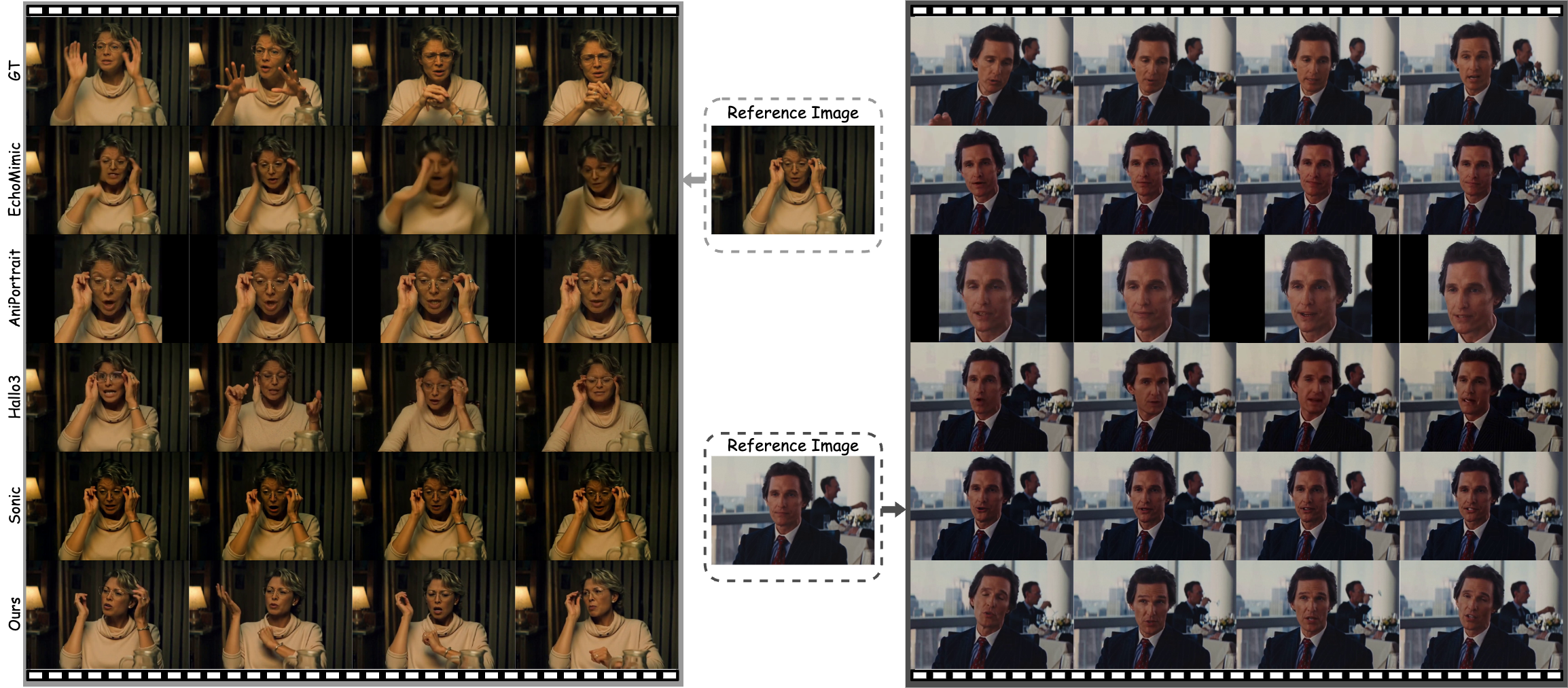}
  \caption{Qualitative comparison on wild talking head dataset.}
  \label{fig:compare_wild}
\end{figure*}

\begin{figure}[h]
  \includegraphics[width=\linewidth]{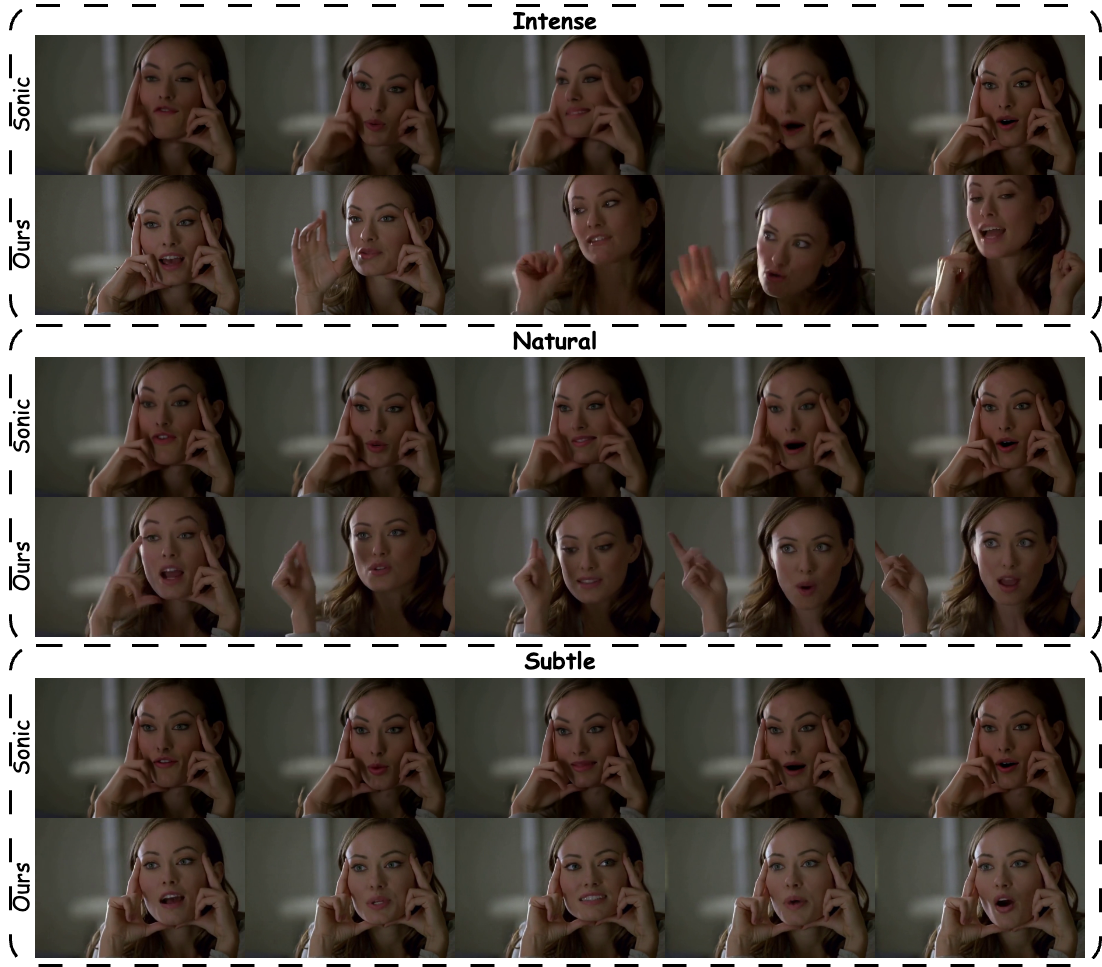}
  \caption{Comparison of Motion Intensity Controller with Sonic.}
  \label{fig:compare_sonic}
\end{figure}

\section{Experiments}

\subsection{Setups}

\noindent {\bf{Implementation Details.}}~ We adopt Wan2.1-I2V-14B \cite{wan2.1} as the foundational model. During the clip-level training stage, we train for approximately 80,000 steps, and during the frame-level training stage, we train for approximately 20,000 steps. Throughout all training phases, both the identity network and the motion network are incorporated into end-to-end training. We employ Flow Matching \cite{lipman2022flow} to train the model, with the entire training conducted on 64 A100 GPUs. The learning rate is set to 1e-4. $\lambda_1$ is set to 1, $\lambda_2$ is set to 0.5, and $\eta$ is set to 0.2. To enhance video generation variability, the reference image, guiding audio and prompt are each set to be independently discarded with a probability of 0.1. 
In the inference stage, we employ the sampling steps of 30, the motion intensity parameter $\omega_l$ and $\omega_b$ are set to neutral value of 0.5, and the CFG \cite{ho2022classifier} of audio is set to 4.5.

\noindent {\bf{Datasets.}}~ The training dataset we use consists of three parts: Hallo3 \cite{cui2024hallo3}, Celebv-HQ \cite{zhu2022celebv}, and data collected from the internet. We utilize InsightFace \cite{deng2019arcface, deng2020retinaface} to exclude videos with a facial confidence score below 0.9 and remove clips \cite{chung2017out} where the speech and mouth motion are not synchronized. This filtering process results in approximately 150,000 clips. We use 50 clips from the HDTF \cite{zhang2021flow} for evaluating the tame talking head generation. Additionally, we evaluate our model on the collected wild talking dataset containing 80 different individuals.

\noindent {\bf{Evaluation Metric and Basedlines.}}~ We employ eight metrics for evaluation. Frechet Inception Distance (\textbf{FID}) \cite{heusel2017gans} and Fréchet Video Distance ({\bf{FVD}}) \cite{unterthiner2019fvd} are used to assess the quality of the generated data. 
\textbf{Sync-C}\cite{chung2017out} and \textbf{Sync-D}\cite{chung2017out} is utilized to measure the synchronization between audio and lip movements.  
The Expression Similarity ({\bf{ES}}) method extracts facial features between video frames \cite{deng2019accurate} and calculates the similarity between these features to evaluate the preservation of identity characteristics. ID consistency (\textbf{IDC}) is achieved by extracting the facial region and computing the DINO \cite{caron2021emerging} similarity metric between frames to measure the consistency of the character's identity features. We utilize SAM \cite{kirillov2023segment} to segment the frame into foreground and background, and separately measure the optical flow scores \cite{teed2020raft} for the foreground and background to evaluate Subject Dynamics (\textbf{SD}) and Background Dynamics (\textbf{BD}), respectively. \textbf{Aesthetic} quality is evaluated using the LAION aesthetic predictor \cite{LAIONAIAestheticPredictor} to assess the artistic and aesthetic value of videos. 

We have selected several state-of-the-art methods to evaluate our approach, all of which have publicly available code or implementations. These methods include the UNet-based approaches Aniportrait \cite{wei2024aniportrait}, EchoMimic \cite{chen2024echomimic} and Sonic \cite{ji2024sonic}, as well as the DiT-based method Hallo3 \cite{cui2024hallo3}. For fair comparison, our method sets the prompt to empty during inference.

\subsection{Results and Analysis}

\noindent {\bf{Comparison on Tame Dataset.}}~ The tame talking head dataset features limited variability in background and character poses, with a primary focus on lip synchronization and facial expression accuracy. Table \ref{tab:comparison} and Figure \ref{fig:compare_hdtf} present the evaluation results. Our method achieves the best scores in FID, FVD, IDC, ES, and Aesthetic score. This success is mainly attributed to our model's ability to generate videos with the most natural and expressive facial expressions, resulting in the highest quality and aesthetically pleasing video outcomes. Additionally, our method achieves the best or second-best results in Sync-C and Sync-D, indicating that our DAVA approach enables the model to learn accurate audio synchronization.

\noindent {\bf{Comparison on Wild Dataset.}}~ Table \ref{tab:comparison} and Figure \ref{fig:compare_wild} present the evaluation results on the wild talking head dataset, which includes significant variations in both foreground and background elements. Previous methods heavily rely on reference images, which limits the naturalness of the generated facial expressions, head movements, and background dynamics. In contrast, our method achieves the best results across all metrics, producing outputs with more natural variations in both foreground and background, improved lip synchronization, and higher overall video quality. This performance is primarily due to our DAVA approach and the identity preservation method focused on facial features. These methods enable our model to better understand the input audio, thereby generating more complex and natural head and background movements while preserving the character's identity features. As a result, our approach better meets the demands of practical application scenarios. 

\begin{table}[ht!]
    \centering
    \small
    \begin{tabular}{ll|c|c|c|c|c}
        \toprule
        \textbf{Level} & \textbf{Method} & \textbf{FVD$\downarrow$} & \textbf{Sync-C$\uparrow$} & \textbf{Sync-D$\downarrow$} & \textbf{IDC$\uparrow$} & \textbf{SD$\uparrow$} \\
        \midrule 
        \multirow{1.8}{*}{subtle} & Sonic & 508.66 & 2.64 & 11.23 & 0.978 & \textbf{8.32} \\
        & Ours & \textbf{496.22} & \textbf{3.11} & \textbf{10.04} & \textbf{0.982} & 8.12 \\
        \midrule 
        \multirow{1.8}{*}{natural} & Sonic & 489.99 & 2.69 & 10.19 & 0.988 & 10.48 \\
        & Ours & \textbf{483.11} & \textbf{3.15} & \textbf{9.69} & \textbf{0.989} & \textbf{13.78}  \\
        \midrule 
        \multirow{1.8}{*}{intense} & Sonic & 522.78 & 2.06 & 12.59 & 0.971 & 12.32  \\
        & Ours & \textbf{501.67} & \textbf{3.09} & \textbf{9.81} & \textbf{0.980} & \textbf{18.14} \\
        \bottomrule
    \end{tabular}
    \caption{\textbf{Comparison of Motion Intensity Controller with Sonic.}}
    \label{tab:comparison_sonic}
\end{table}

\begin{figure}[h]
  \includegraphics[width=\linewidth]{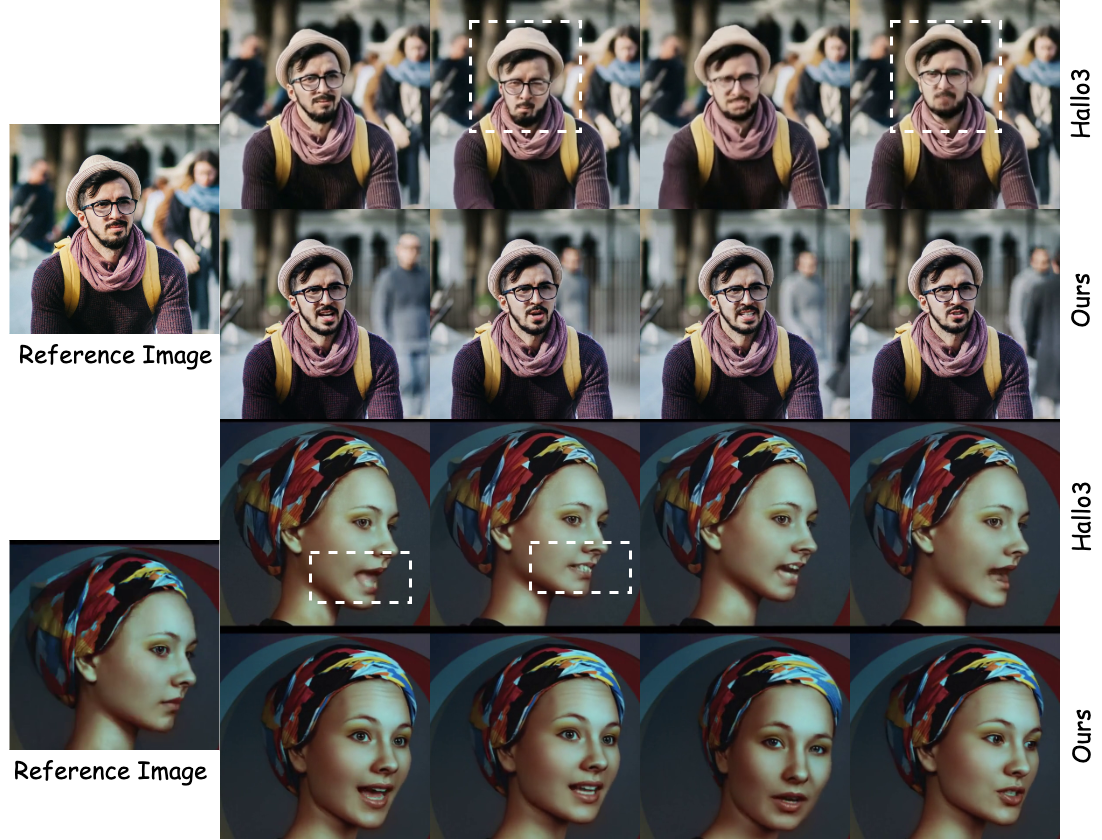}
  \caption{Comparison of Visualization Results with Hallo3.}
  \label{fig:compare_hallo3}
\end{figure}

\noindent {\bf{Comparison of Motion Intensity Controller with Sonic.}}~ In our comparative study, Sonic exhibits a similar ability to control motion intensity, allowing users to regulate the expressiveness and head movement through an input parameter $\beta$. We conducted comparative experiments by categorizing the motion intensity into three levels: subtle ($\beta$=0.5, $\omega_l$=0.1, $\omega_b$=0.1), natural ($\beta$=1.0, $\omega_l$=0.5, $\omega_b$=0.5) and intense ($\beta$=2.0, $\omega_l$=1.0, $\omega_b$=1.0). The experimental results are presented in Table \ref{tab:comparison_sonic} and Figure \ref{fig:compare_sonic}. At the natural and subtle levels, both our method and Sonic demonstrate excellent control over motion intensity while maintaining lip synchronization. However, in scenarios involving intense movements, our method achieves superior results. This is because our limb control approach focuses on the entire body movement, including the head, whereas Sonic only considers head movements. Consequently, our method exhibits a more competitive ability in representing the full range of human motion.

\begin{figure}[h]
  \includegraphics[width=\linewidth]{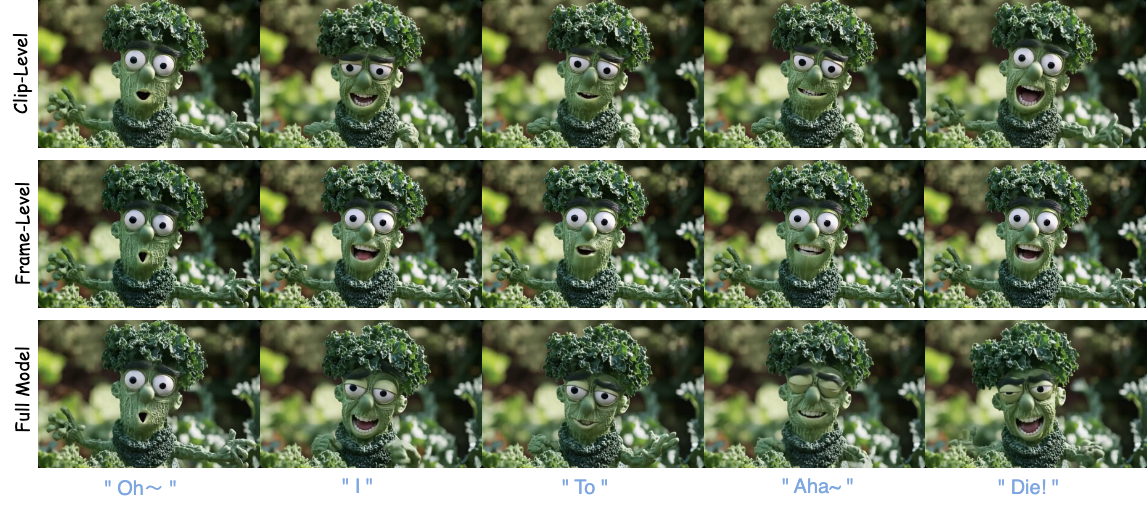}
  \caption{Ablation on DAVA.}
  \label{fig:ablation_1}
\end{figure}

\noindent {\bf{Comparison of Visualization Results with Hallo3.}}~ We present additional visualization comparisons with Hallo3 in Figure \ref{fig:compare_hallo3}, which is a DiT-based method for generating wild talking head videos. Our approach demonstrates more realistic results. For instance, the outputs of Hallo3 exhibit noticeable distortions and artifacts on the person's face and lips,  as well as unrealistic background movements in the top row of \ref{fig:compare_hallo3}, and relatively stiff head movements in the bottom row of \ref{fig:compare_hallo3}. In contrast, our results showcase more authentic expressions, head movements, and background dynamics. These improvements can be attributed to our focus on facial knowledge learning, which enhances the identity features of the person, and the DAVA method, which strengthens the learning of lip synchronization. 

\noindent {\bf{User Studies.}}~ To further validate the effectiveness of our proposed method, we conducted a subjective evaluation on the Wild Talking Head dataset. Each participant assessed four critical dimensions: Lip Synchronization (\textbf{LS}), Video Quality (\textbf{VQ}), Identity Preservation (\textbf{IP}), and Motion Diversity (\textbf{MD}). A total of 24 participants rated each aspect on a scale from 0 to 10. As shown in Table \ref{tab:userstudy}, the scores demonstrate that FantasyTalking outperforms baseline methods across all evaluated dimensions, exhibiting particularly notable improvements in motion diversity. This comprehensive evaluation highlights the superiority of our approach in generating realistic and diverse talking head animations while maintaining consistent identity representation and high visual fidelity.

\begin{table}[H]
    \centering
    \small
    \begin{tabular}{>{\raggedright\arraybackslash}p{1.8cm}|
                    >{\centering\arraybackslash}p{1.08cm}|
                    >{\centering\arraybackslash}p{1.08cm}|
                    >{\centering\arraybackslash}p{1.08cm}|
                    >{\centering\arraybackslash}p{1.08cm}}
        \toprule
        \textbf{Method} & \textbf{LS} & \textbf{VQ} & \textbf{IP} & \textbf{MD} \\
        \midrule 
        Aniportrait & 8.18 & 6.78 & 7.82 & 5.28 \\
        EchoMimic & 8.22 & 6.31 & 7.05 & 4.40 \\
        Sonic & 9.07 & 8.17 & 8.13 & 6.25 \\
        Hallo3 & 8.93 & 7.89 & 7.82 & 6.44 \\
        FantasyTalking & \textbf{9.45} & \textbf{9.18} & \textbf{8.44} & \textbf{9.81}   \\
        \bottomrule
    \end{tabular}
    \caption{\textbf{User Study results.}}
    \label{tab:userstudy}
\end{table}

\section{Ablation Studies and Discussion} 

\begin{table}[H]
    \centering
    \small
    \begin{tabular}{>{\raggedright\arraybackslash}p{1.8cm}|
                    >{\centering\arraybackslash}p{0.8cm}|
                    >{\centering\arraybackslash}p{1.08cm}|
                    >{\centering\arraybackslash}p{1.08cm}|
                    >{\centering\arraybackslash}p{0.7cm}|
                    >{\centering\arraybackslash}p{0.7cm}}
        \toprule
        \textbf{Method} & \textbf{FVD$\downarrow$} & \textbf{Sync-C$\uparrow$} & \textbf{Sync-D$\downarrow$} & \textbf{IDC$\uparrow$} & \textbf{SD$\uparrow$} \\
        \midrule 
        Clip-Level & 492.85 & 1.98 & 11.21 & 0.986 & 13.66 \\
        Frame-Level & 534.39 & \textbf{3.54} & \textbf{9.02} & 0.987 & 8.22 \\
        w/o Identity & 510.62 & 3.06 & 10.15 & 0.945 & 12.96 \\
        FantasyTalking & \textbf{483.11} & 3.15 & 9.69 & \textbf{0.989} & \textbf{13.78} \\
        \bottomrule
    \end{tabular}
    \caption{\textbf{Ablation studies on DAVA and Identity Preservation in Wild Dataset.}}
    \label{tab:ablation}
\end{table}

\begin{figure}[h]
  \includegraphics[width=\linewidth]{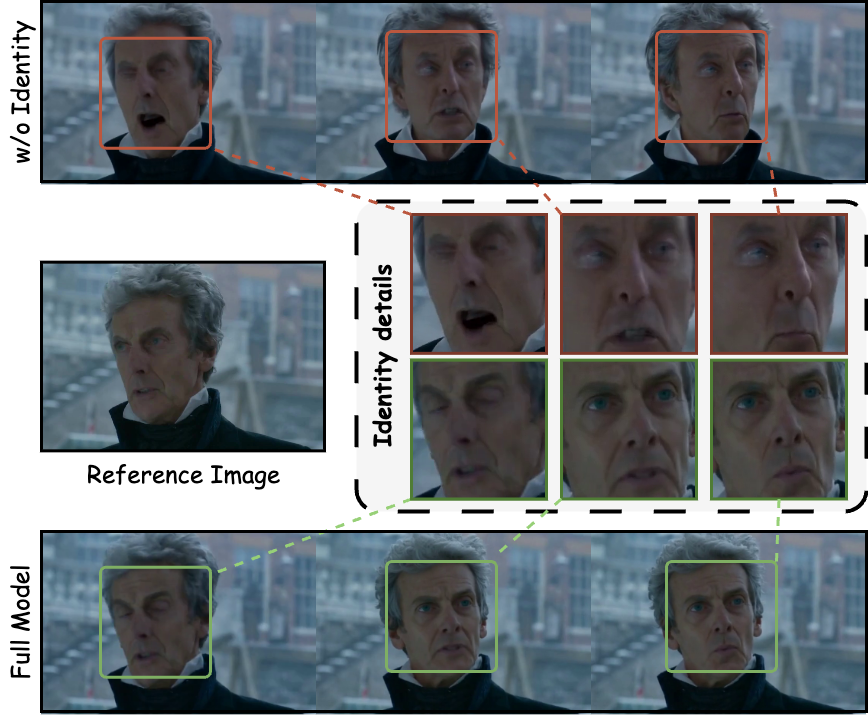}
  \caption{Ablation on Identity Preservation.}
  \label{fig:ablation_identity}
\end{figure}

\noindent {\bf{Ablation on DAVA. }}~ To validate the effectiveness of our DAVA method, we performed experiments using audio-visual alignment at clip level and only at frame level for training. The results, as presented in Table \ref{tab:ablation} and illustrated in Figure \ref{fig:ablation_1}. Training with only clip-level alignment leads to a significant decline in the Sync-C metric. This indicates that relying solely on clip-level alignment is insufficient to learn the precise correspondence between audio and lip movements. However, training with only frame-level alignment, while demonstrating strong lip-sync capabilities, noticeably limits the dynamic nature of facial expressions and subject movements. In contrast, our proposed DAVA method effectively combines the advantages of both clip-level and frame-level alignments, which achieves precise audio-to-lip synchronization while enhancing the vividness of character animations and background dynamics.

\noindent {\bf{Ablation on Identity Preservation.}}~ The results presented in Table \ref{tab:ablation} underscore the importance of identity preservation in our model. Without identity preservation, IDC significantly decreases, which implies that the model's ability to maintain the character's identity features is greatly reduced, leading to a decline in video quality. As shown in Figure \ref{fig:ablation_identity}, the absence of identity preservation lead to artifacts and distortions in the facial features. In contrast, our proposed identity preservation method, which incorporates focused facial knowledge learning, enhances the model's ability to maintain the character's identity while preserving lip synchronization and rich motion capabilities. This leads to improved identity retention and overall video quality.

\begin{figure}[h]
  \includegraphics[width=\linewidth]{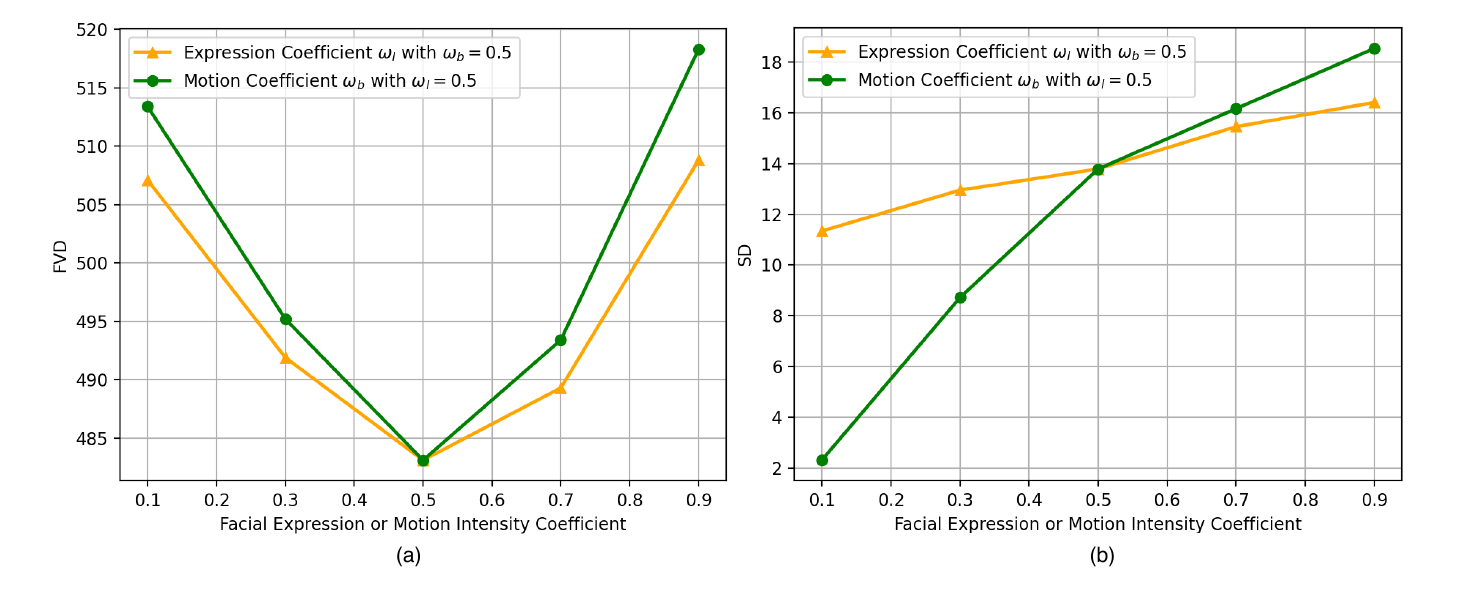}
  \caption{Ablation on Motion Intensity Modulation Network.}
  \label{fig:ablation_motion}
\end{figure}

\noindent {\bf{Ablation on Motion Intensity Modulation Network.}}~ Figure \ref{fig:ablation_motion} illustrates the quantitative results of adjusting the motion intensity coefficient $\omega_l$ and $\omega_b$ on FVD and SD. When one parameter is varied, the other is fixed at a neutral value of 0.5. As shown in Figure \ref{fig:ablation_motion} (a), the results with natural motion intensity ($\omega_l=0.5$, $\omega_b=0.5$) achieve the best FVD scores. This suggests that facial and body motion intensities that are either too high or too low tend to produce visual representations that deviate from realistic scenarios, which result in less authentic visual representations. Figure \ref{fig:ablation_motion}(b) demonstrates that as the $\omega_l$ or $\omega_b$ parameters increase, the subject dynamic score becomes significantly more pronounced. This highlights the effectiveness of our motion control mechanism, which provides users with a tool for explicitly controlling the speaking motion intensity. 

\noindent {\bf{Limitations and Future Works.}}~ Despite the significant progress achieved by our method, especially in the scenario of wild talking head video generation, due to the iterative sampling process required by the diffusion model during inference to achieve optimal results, the overall runtime can be relatively slow. Investigating acceleration strategies would facilitate its use in scenarios with higher real-time requirements, such as live streaming and interactive real-time applications. Furthermore, investigating interactive portrait dialogue solutions with real-time feedback based on audio-driven talking head generation can broaden applications in realistic digital human avatar scenarios.

\section{Conclusions}

In this paper, we introduce FantasyTalking, a novel audio-driven portrait animation technique. By employing a dual-stage audio-visual alignment training process, our method effectively captures the relationship between audio signals and lip movements, facial expressions, as well as body motions. To enhance identity consistency within the generated videos, we propose a facial-focused approach to retain facial features accurately. Additionally, a motion network is utilized to control the magnitude of facial expressions and body movements, ensuring natural and varied animations. Both qualitative and quantitative experiments demonstrate that FantasyTalking outperforms existing SOTA methods in several key aspects, including video quality, motion diversity, and identity consistency.

\newpage


\bibliographystyle{ACM-Reference-Format}
\bibliography{arxiv}

\end{document}